\documentclass[10pt,twocolumn,letterpaper]{article}

\usepackage{algorithm}
\usepackage{algorithmic}
\usepackage{tabularx}
\usepackage{makecell}
\usepackage{multirow}
\usepackage{booktabs}
\usepackage{bm}
\usepackage{color}
\usepackage{amssymb} 
\usepackage{amsmath}

\usepackage[pagenumbers]{cvpr}      

\definecolor{cvprblue}{rgb}{0.21,0.49,0.74}
\usepackage[pagebackref,breaklinks,colorlinks,allcolors=cvprblue]{hyperref}

\def\confName{CVPR}
\def\confYear{2025}

\begin{document}
\title{DeltaEdit: Exploring Text-free Training for Text-Driven Image Manipulation}

\author{
Yueming Lyu$^{1,2}$, Tianwei Lin$^{3}$, Fu Li$^{3}$, Dongliang He$^{3}$, Jing Dong$^{2*}$, Tieniu Tan$^{2,4}$\\
$^{1}$ School of Artificial Intelligence, University of Chinese Academy of Sciences\\
$^{2}$ CRIPAC, Institute of Automation, Chinese Academy of Sciences\\
$^{3}$ VIS, Baidu Inc.\\
$^{4}$ Nanjing University\\
{\tt\small yueming.lv@cripac.ia.ac.cn, jdong@nlpr.ia.ac.cn}}

\twocolumn[{
\renewcommand\twocolumn[1][]{#1}
\maketitle

\vspace{-0.9cm}
\begin{center}
\captionsetup{type=figure}
\includegraphics[width=0.8\linewidth]{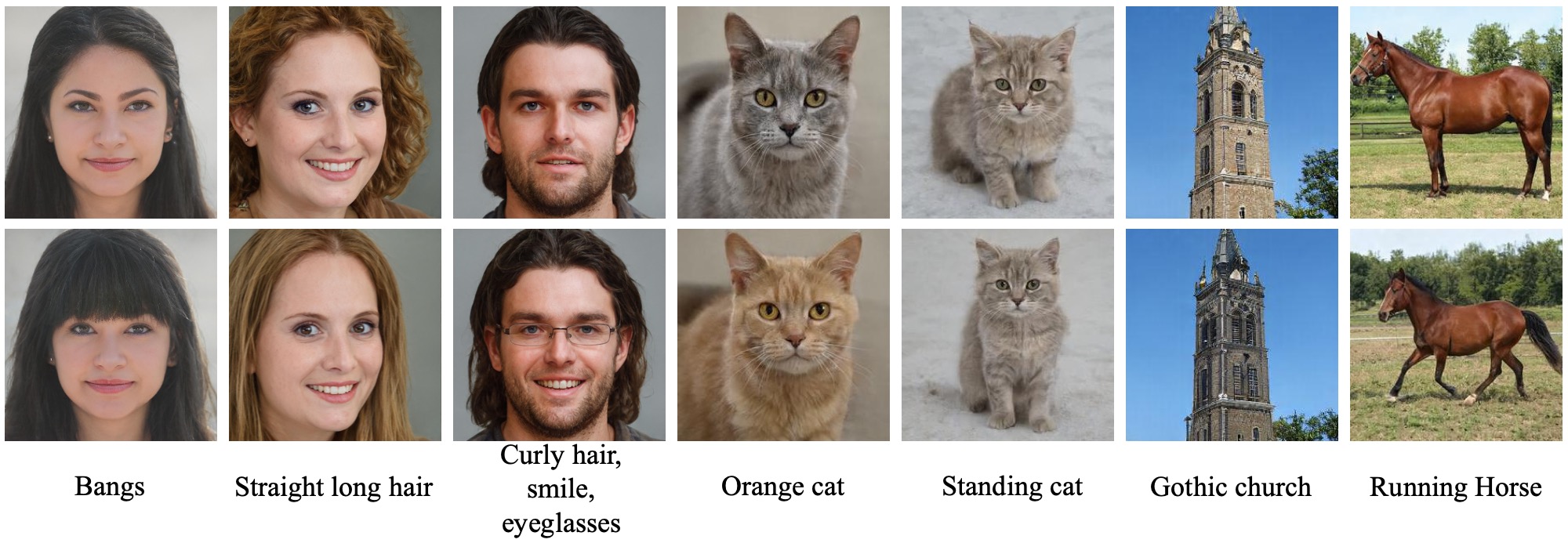}
\vspace{-0.3cm}
\captionof{figure}{Examples of text-driven manipulations of multiple StyleGAN imagery categories induced by our DeltaEdit with text-free training.}
\label{fig:shouye}
\end{center}
}]
\renewcommand{\thefootnote}{\fnsymbol{footnote}} 
\footnotetext[1]{Corresponding author.} 

\begin{abstract}
Text-driven image manipulation remains challenging in training or inference flexibility. Conditional generative models depend heavily on expensive annotated training data. Meanwhile, recent frameworks, which leverage pre-trained vision-language models, are limited by either per text-prompt optimization or inference-time hyper-parameters tuning. In this work, we propose a novel framework named \textit{DeltaEdit} to address these problems. Our key idea is to investigate and identify a space, namely delta image and text space that has well-aligned distribution between CLIP visual feature differences of two images and CLIP textual embedding differences of source and target texts. Based on the CLIP delta space, the DeltaEdit network is designed to map the CLIP visual features differences to the editing directions of StyleGAN at training phase. Then, in inference phase, DeltaEdit predicts the StyleGAN's editing directions from the differences of the CLIP textual features. In this way, DeltaEdit is trained in a text-free manner. Once trained, it can well generalize to various text prompts for zero-shot inference without bells and whistles. 
\end{abstract}
    
\section{Introduction}
\label{sec:intro}

\hspace*{1em} Text-driven image manipulation has aroused widespread research interests in both academic and industrial communities given its significance for real-world applications. It aims at editing the content of images according to user-provided natural language descriptions while preserving text-irrelevant content unchanged.

Existing approaches~\cite{dong2017semantic,nam2018text,liu2020describe,li2020manigan,xia2021tedigan} typically train conditional generative models from scratch with a large amount of manually annotated image-text pairs. This procedure requires expensive labor annotation, which obstacles the training flexibility. Recently, some CLIP+StyleGAN approaches~\cite{patashnik2021styleclip,liu2021fusedream,kocasari2022stylemc,wei2021hairclip,zhou2021lafite} have been proposed to perform text-driven image manipulation by utilizing the CLIP's remarkable semantic representation of image-text modalities, and well-trained StyleGAN's high-quality generation power. Given one text prompt, these methods either leverage iterative optimization~\cite{xia2021tedigan,patashnik2021styleclip,liu2021fusedream} or learn the corresponding mapping network~\cite{patashnik2021styleclip,gal2021stylegan,xu2022predict}, or require to tune the complex hyper-parameters online to discover the specific editing direction~\cite{patashnik2021styleclip}.
Namely, for different text prompts, they must implement different optimization processes, which is not flexible during training or inference, and cannot well generalize to any other unseen text. 

We consider that the key to alleviate these problems is to precisely build the relationships between the text feature space and the StyleGAN's~\cite{Karras2019stylegan2} latent visual space within one model. Manually collecting relevant textual data to train one model is a possible way~\cite{wei2021hairclip,zhu2022one}, but it could only learn the sub-relationships and lead to limited generalization. Therefore, it is challenging but desired to explore how to construct the full mapping between two feature spaces without any textual supervision to restrict the generalization ability.

To achieve this goal, we figure out a \emph{text-free training} paradigm, where a text-driven image manipulation model is trained on easily collected image data instead of textual data, but the model can be well generalized to input of text prompts for zero-shot inference. The key of approaching this is to identify a well semantically aligned feature space for image and text modalities. Within such an aligned space, image features can be utilized as pseudo text conditions to achieve text-free training. 

\begin{figure}[t]
	\begin{center}
		\includegraphics[width=0.8\linewidth]{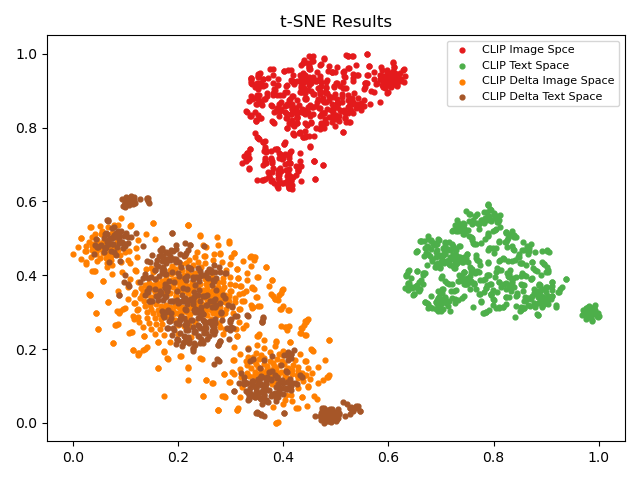}
	\end{center}
	\vspace{-3ex}
	\caption{Feature space analysis on the MultiModal-CelebA-HQ~\cite{xia2021tedigan} dataset. Paired CLIP image-text features (marked in red and green) and paired CLIP delta image-text features (marked in orange and brown) are visualized in 2D using t-SNE visualization.}
	\label{fig:tsne}
\end{figure}

In this paper, we first approach text-free training by a naïve solution, which naturally uses the CLIP image features as pseudo text conditions. However, it fails to perform well-disentangled image manipulation with user-provided text descriptions, which mainly because there is still a modality gap between the image and text feature spaces~\cite{liang2022mind} (illustrated in Figure \ref{fig:tsne}). Notably, it is the direction of CLIP features that is semantically meaningful as all features are normalized to a unit norm \cite{patashnik2021styleclip,zhou2021lafite}, and CLIP feature differences of paired visual-textual data both mean similar semantic changes. Though gap exists between features of each individual modality, we have demonstrated that the CLIP feature space for image-text differences are more consistent and semantically better aligned (shown in Figure \ref{fig:tsne}). We term this feature space as the \emph{CLIP Delta} space. Potentially, constructing pseudo text conditions in the better aligned CLIP delta space will enable more fine-grained and better disentangled edits for the text-free trained manipulation model. 

Based on the above analysis, we propose our DeltaEdit framework. Specifically, we learn a coarse-to-fine mapping network from the CLIP image feature differences of two images to their editing direction at the $\mathcal{S}$ style space, then directly applying the learned mapping onto the differences of two texts' CLIP features could obtain the corresponding editing direction at $\mathcal{S}$ space. The predicted editing direction shall then be able to change image attributes from one text description to the other text description. As shown in Figure~\ref{fig:shouye}, we also demonstrate DeltaEdit can well generalize to various target text-prompts without bells and whistles.

To summarize, our main contributions are as follows:
\begin{itemize}
    \item We push the frontiers of text-free training for image manipulation, which largely eliminates the training or inference inflexibility and poor generalization suffered by previous methods.
    \item We investigate and identify the CLIP delta image-text feature space is better semantically aligned than the original CLIP space. Furthermore, we propose our DeltaEdit framework to learn the mapping from image-feature change to StyleGAN's $\mathcal{S}$ style space change and seamlessly generalize to predict editing direction from text-feature change.
    \item Extensive experiments on a variety of datasets, including FFHQ and LSUN, verify the effectiveness and efficiency of our method. Results show DeltaEdit can well generalize to various edits conditioned on different target text-prompts without bells and whistles.
\end{itemize}
\section{Related Work}

\subsection{Vision-Language Representations}

\hspace*{1em} Learning generic vision-language representations is of great importance in the multi-modal field. Following the success of BERT~\cite{devlin2018bert}, many large scale pre-trained vision-language models~\cite{li2020oscar,lu2019vilbert,su2019vl,tan2019lxmert,zhang2021vinvl} are proposed. A recent development, CLIP~\cite{radford2021learning}, is extremely powerful for joint vision-language learning. It is trained on 400 million (image, text) pairs and learns a joint multi-modal semantic space for both images and texts. Benefit from the excellent image/text representation ability, it has been widely used in various directions, such as domain adaptation~\cite{ge2022domain}, image segmentation~\cite{zhou2021denseclip,rao2021denseclip}, and image generation~\cite{gal2021stylegan,zhu2021mind,xu2022predict,liu2021fusedream,li2022stylet2i,sun2022anyface}. 

\subsection{Text-Guided Generation and Manipulation}

\hspace*{1em} The goal of text-guided image generation~\cite{reed2016generative,cheng2020rifegan,sun2021multi,lyu2021sogan,han2017stackgan,zhang2018stackgan,xu2018attngan} is to generate realistic images according to given text descriptions. 
As a pioneering work, Reed \emph{et al.}~\cite{reed2016generative} embed the text features as the conditional input for a GAN-based one-stage architecture. 
After that, StackGAN~\cite{han2017stackgan} and StackGAN++~\cite{zhang2018stackgan} introduce multi-stage frameworks to further improve the generation quality. Recently, some works~\cite{stap2020conditional,sun2022anyface,xia2021tedigan,zhou2021lafite} based on StyleGAN~\cite{karras2019style,Karras2019stylegan2,karras2021alias} have been proposed to improve the quality of generated images. 

Compared with generating, text-guided image manipulation~\cite{dong2017semantic,nam2018text,liu2020describe,li2020manigan,xia2021tedigan,xu2022predict,kim2022diffusionclip} aims to manipulate the input images by using texts describing desired visual attributes (\emph{e.g.}, gender, age). For example, ManiGAN~\cite{li2020manigan} proposes affine combination module (ACM) and detail correction module (DCM) to generate new attributes matching the given text.  
Recently, some works~\cite{stap2020conditional,xia2021tedigan} have adopted StyleGAN as their backbone to perform image manipulation tasks. For TediGAN~\cite{xia2021tedigan}, it aligns two modalities in the latent space of pre-trained StyleGAN by the proposed visual-linguistic similarity module. More recently, StyleCLIP~\cite{patashnik2021styleclip} combines the generation power of StyleGAN and the image-text representation ability of CLIP~\cite{radford2021learning} to discover manipulation direction. 
They outline three approaches, namely Latent Optimization, Latent Mapper and Global Directions, which are denoted as StyleCLIP-OP, StyleCLIP-LM, and StyleCLIP-GD in this paper. 
The StyleCLIP-OP and StyleCLIP-LM are per-prompt training methods, which require optimizing the latent code or training a separate model for each text prompt. The third approach, StyleCLIP-GD, is a per-prompt fine-tuning method. It first finds global directions of semantic change in StyleGAN's style space $\mathcal{S}$ by pre-defined relevance matrix of each channel in $\mathcal{S}$ to the image-space change. During inference, it needs to manually tune hyper-parameters to discover the fine-grained directions for each specific text prompt. In addition, HairCLIP~\cite{wei2021hairclip} focuses more on the hair manipulation with the help of CLIP. FFCLIP~\cite{zhu2022one} collects 44 text prompts for face images to learn the facial attributes manipulation.

\section{Method}
\hspace*{1em} Our method aims to predict editing directions in StyleGAN $\mathcal{S}$ space conditioned on corresponding embeddings in CLIP image space, without any text supervision. 

Randomly given two images from the training image dataset, one as the source image $I_1$ and the other as the target image $I_2$. Then the CLIP image encoder is used to extract their CLIP image embeddings, $i_1$ and $i_2$, and the pre-trained StyleGAN inversion model~\cite{tov2021designing} is adopted as an encoder to extract the latent codes $s_1$ and $s_2$ in the $\mathcal{S}$ space, which has been proven to be more disentangled than other intermediate feature spaces~\cite{wu2021stylespace}. Taking the extracted codes, we predict the manipulation direction $\Delta s = s_2 - s_1$.

In Section~\ref{sec:solution1}, we start by proposing a naïve solution to construct pseudo text conditions directly with image embeddings, based on the alignment property of the joint CLIP's image-text space. However, the resulting method cannot perform well when editing images, and we further propose a new text-free training framework, called DeltaEdit, in Section~\ref{solution2}, which alleviates the problem of the naïve solution and performs accurate and disentangled editing in the proposed CLIP delta space.

\subsection{A Naïve Solution to Text-free Training}
\label{sec:solution1}
\begin{figure}[t]
	\begin{center}
		\includegraphics[width=1\linewidth]{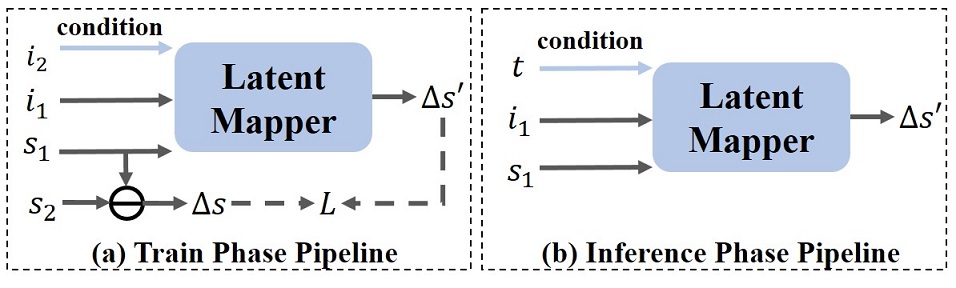}
	\end{center}
	\vspace{-2ex}
	\caption{Illustration of the naïve solution to text-free training.}
	\label{fig:naive_framework}
\end{figure}

\hspace*{1em} As illustrated in Figure~\ref{fig:naive_framework}, in the training phase, the source embeddings, $i_1$ and $s_1$, are taken as the input. The condition is from a target CLIP image embedding $i_2$ instead of a target CLIP text embedding. Then, the source embeddings and the image condition are sent to a latent mapper network to predict the manipulation direction $\Delta s'$:
\begin{equation}
\Delta s' = LatentMapper(s_1, i_1, i_2)\text{.}
\end{equation}
During inference, we can predict the editing direction $\Delta s'$ as:
\begin{equation}
\Delta s' = LatentMapper(s_1, i_1, t)\text{,}
\end{equation}
where $t$ is a CLIP text embedding constructed from a target text prompt.
\begin{figure}[t]
	\begin{center}
		\includegraphics[width=1\linewidth]{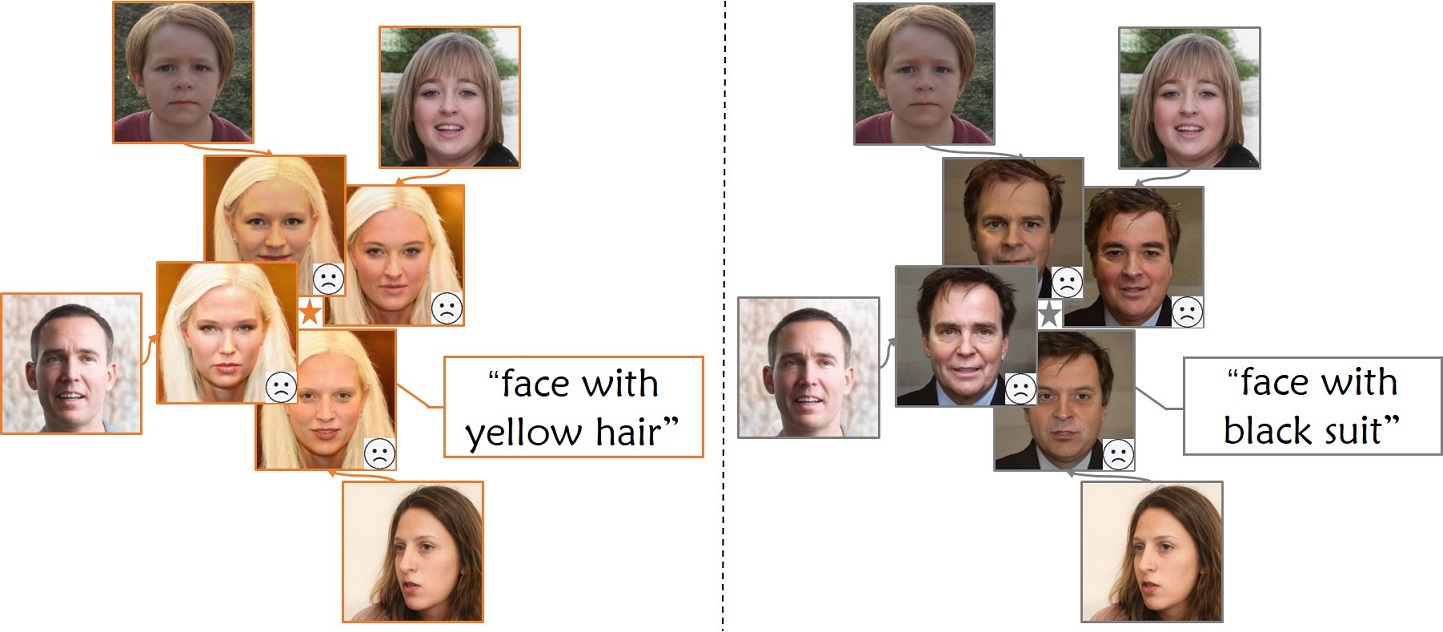}
	\end{center}
	\vspace{-2ex}
	\caption{The editing results of the naïve solution. Take different source images as input, the method fits them all to an average face corresponding to the text-related attributes, by directly replace the condition from image feature $i$ to text feature $t$.}
	\label{fig:naive}
\end{figure}
\begin{figure*}[t]
	\begin{center}
		\includegraphics[width=0.9\linewidth]{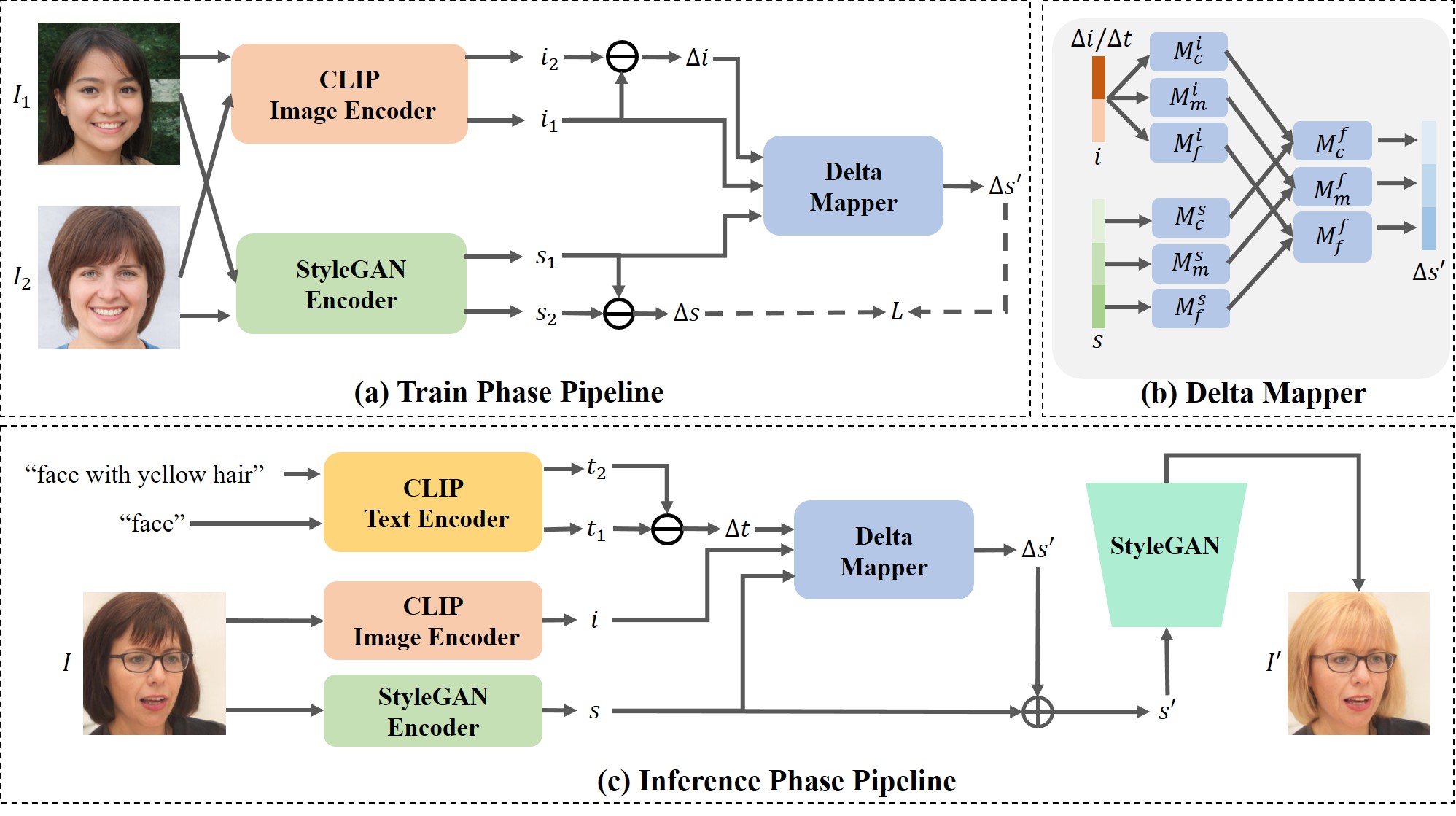}
	\end{center}
	\vspace{-3ex}
	\caption{The overall framework of the proposed DeltaEdit. 
	(a) In the text-free training phase, we extract embeddings of two randomly selected images on CLIP image space and StyleGAN $\mathcal{S}$ space. Then we feed $i_1$, $s_1$ and $\Delta i$ into Delta Mapper to predict editing direction $\Delta s'$, which is supervised by $\Delta s$. 	(b) The detailed architecture of the Delta Mapper, which achieves coarse-to-fine manipulation in three levels. 	(c) In inference phase, based on the co-linearity between $\Delta i$ and $\Delta t$ in CLIP joint space, DeltaEdit can achieve text-driven image manipulation by taking two text prompts (denoting the source and desired target) as input.}
	\label{fig:framework}
\end{figure*}

However, as shown in Figure~\ref{fig:naive}, given different source images, the naïve method maps them all to average faces corresponding to the text-related attributes, leading to inaccuate manipulation. 
We doubt that this phenomenon occurs because there is still a modality gap of the CLIP visual-linguistic modalities, even though the cosine similarity between paired image-text features are maximized to make the image-text space aligned in CLIP. To verify it, we perform t-SNE visualization on the MM-CelebA-HQ dataset~\cite{xia2021tedigan} in Figure~\ref{fig:tsne}, which shows that the CLIP image space and text space are actually not that close to each other. A recent work~\cite{liang2022mind} has also demonstrated that the joint multi-modal space in CLIP is not well-aligned caused by model initialization and contrastive representation learning. 

After further practice, it is encouraging that we have identified a better semantically aligned multi-modal feature space, namely the CLIP delta image-text space, which is extremely close to each other, making it possible to approximate the text conditions with image embeddings in the proposed delta space. Additionally, there is an advantage of leveraging the text embeddings change as conditions, such as ``face'' as the source text and ``face with yellow hair'' as the target text. That is, the importance of words that indicate the desired attributes (like ``yellow hair'') can be enhanced, which alleviates the problem that CLIP is not sensitive to fine-grained or complex words~\cite{radford2021learning,saharia2022photorealistic}.

\subsection{DeltaEdit}
\label{solution2}
\hspace*{1em} In our DeltaEdit framework, we extract delta image-text features, $\Delta i = i_2 - i_1$ and $\Delta t = t_2 - t_1$, as the condition vectors in the training phase and inference phase, respectively. 

As shown in Figure~\ref{fig:framework}, during training, we take the extracted codes to get CLIP image space direction $\Delta i = i_2 - i_1$ and StyleGAN $\mathcal{S}$ space direction $\Delta s=s_2 -s_1$. Then, we propose a latent mapper network in delta space, called Delta Mapper, to predict the editing direction as:
\begin{equation}
\Delta s' = DeltaMapper(s_1, i_1, \Delta i)\text{,}
\end{equation}
where $s_1$ and $i_1$ are used as the input of Delta Mapper to provide specialized information for the source image.

\textbf{Delta Mapper.} The architecture of the Delta Mapper is illustrated in Figure~\ref{fig:framework} (b). Since StyleGAN has the property that different layers correspond to different semantic levels~\cite{karras2019style,xia2021tedigan}, it is common to divide these layers into different levels and implement coarse-to-fine manipulation within each level. Following \cite{patashnik2021styleclip,wei2021hairclip}, we adopt three levels of sub-modules~(coarse, medium, and fine) for each designed module. Each sub-module contains several fully-connected layers. In this way, with different levels of the source latent code $s_1$, we propose \emph{Style Module} to obtain coarse-to-fine intermediate features $(e^s_c, e^s_m, e^s_f)$, where subscripts stand for coarse-to-fine level and superscripts stand for the Style Module. Then, we concatenate the $\Delta i$ and $i_1$ as input, and propose \emph{Condition Module} to learn coarse, medium and fine embeddings $(e^i_c, e^i_m, e^i_f)$ separately, which have same dimensions as $(e^s_c, e^s_m, e^s_f)$. In the final step, we fuse generated coarse-to-fine features using proposed \emph{Fusion Module} with three sub-modules ($M^f_c(\cdot,\cdot), M^f_m(\cdot,\cdot), M^f_f(\cdot,\cdot)$) and predict editing direction as:
\begin{equation}
    \begin{aligned}
    \Delta s^{\prime} = &(M^f_c(e^i_c, e^s_c), M^f_m(e^i_m, e^s_m), M^f_f(e^i_f, e^s_f)). 
    \end{aligned}
\end{equation}

To train the proposed Delta Mapper, our full objective function contains two losses, which can be denoted as:

\begin{equation}
    \begin{aligned}
       \mathcal{L}\!=\!\mathcal{L}_{rec}\!+\!\mathcal{L}_{sim}\!=\! \|\Delta s^{\prime}\!-\!\Delta s\|_2 +\!1\!- cos(\Delta s^{\prime},\Delta s),
    \end{aligned}
\end{equation}
where L-2 distance reconstruction loss is utilized to add supervision for learning the editing direction $\Delta s^{\prime}$ and cosine similarity loss is introduced to explicitly encourage the network to minimize the cosine distance between the predicted embedding direction $\Delta s^{\prime}$ and $\Delta s$ in the $\mathcal{S}$ space.

\textbf{Text-driven Image Manipulation.}
During inference, based on the well-aligned multi-modal delta feature space, we can achieve text-driven image manipulation with the trained Delta Mapper. As shown in Figure~\ref{fig:framework} (c), given the source image $I$, we first extract its CLIP image embedding $i$ and StyleGAN $\mathcal{S}$ space embedding $s$. Then, we construct $\Delta t$ with source and target text prompt, and we can predict the editing direction $\Delta s'$ as:
\begin{equation}
\Delta s' = DeltaMapper(s, i, \Delta t),
\end{equation}
where the editing direction is  subsequently used to generate edited latent embedding $s' = s + \Delta s'$. In the final step, we can generate the synthesized image $I'$ with StyleGAN based on $s'$. 

\textbf{Text Prompts.}
In inference phase pipeline, the crucial issue is how to construct $\Delta t = t_2 - t_1$. 
To establish the correspondence with the training phase, the source text and target text should be specific text descriptions of two different images. However, the DeltaEdit network will generate images towards the direction of the target text and the opposite direction of the source text in this manner. 
Therefore, to remove the reverse effect of the source text prompt containing attributes, we naturally place all the user-described attributes in the target text prompts.
For human portrait images, if the user intends to add ``smile'' and ``eyeglasses'' to a face, then our DeltaEdit uses ``face with smile and eyeglasses'' as the target text and ``face'' as the source text. 
This text prompt setting is also applicable to other manipulation domains.
In experiments, we further verify that it enables accurate image manipulation.

\textbf{Disentanglement.}
To improve the disentanglement, we further optimize the obtained editing direction $\Delta s^{\prime}$ using the pre-computed relevance matrix $R_s$, which records how CLIP image embedding changes when modifying each dimension in $\mathcal{S}$ space ~\cite{patashnik2021styleclip}. We can set some channels of $\Delta s^{\prime}$ as zero if the channels have a low correlation to the target text according to $R_s$.
\section{Experiments}
\subsection{Implementation Details}
\hspace*{1em} To verify the effectiveness and generalization of the proposed method, we conduct extensive experiments on a diverse set of challenging domains. For the facial domain, we randomly choose 58,000 images from FFHQ~\cite{karras2019style} dataset and sample 200,000 fake images from the $z$ space in StyleGAN for training; we use the remained 12,000 FFHQ~\cite{karras2019style} images for evaluation.
We additionally provide results on the LSUN~\cite{yu2015lsun} Cat, Church and Horse datasets. Note that all real images are inverted by e4e encoder~\cite{tov2021designing}, and all generated images are obtained using pre-trained StyleGAN2~\cite{Karras2019stylegan2} generators. The proposed method is trained on 1 NVIDIA Tesla P40 GPU. During training, we set batch size as 64 and adopt ADAM~\cite{kingma2014adam} optimizer with $\beta_1=0.9$, $\beta_2=0.999$ and a constant learning rate of 0.5.

\subsection{Qualitative Evaluation}
\hspace*{1em} We perform multiple qualitative evaluations with different text descriptions on several challenging datasets.
\emph{More results are shown in Supp.}

\begin{figure*}[t]
	\begin{center}
		\includegraphics[width=0.9\linewidth]{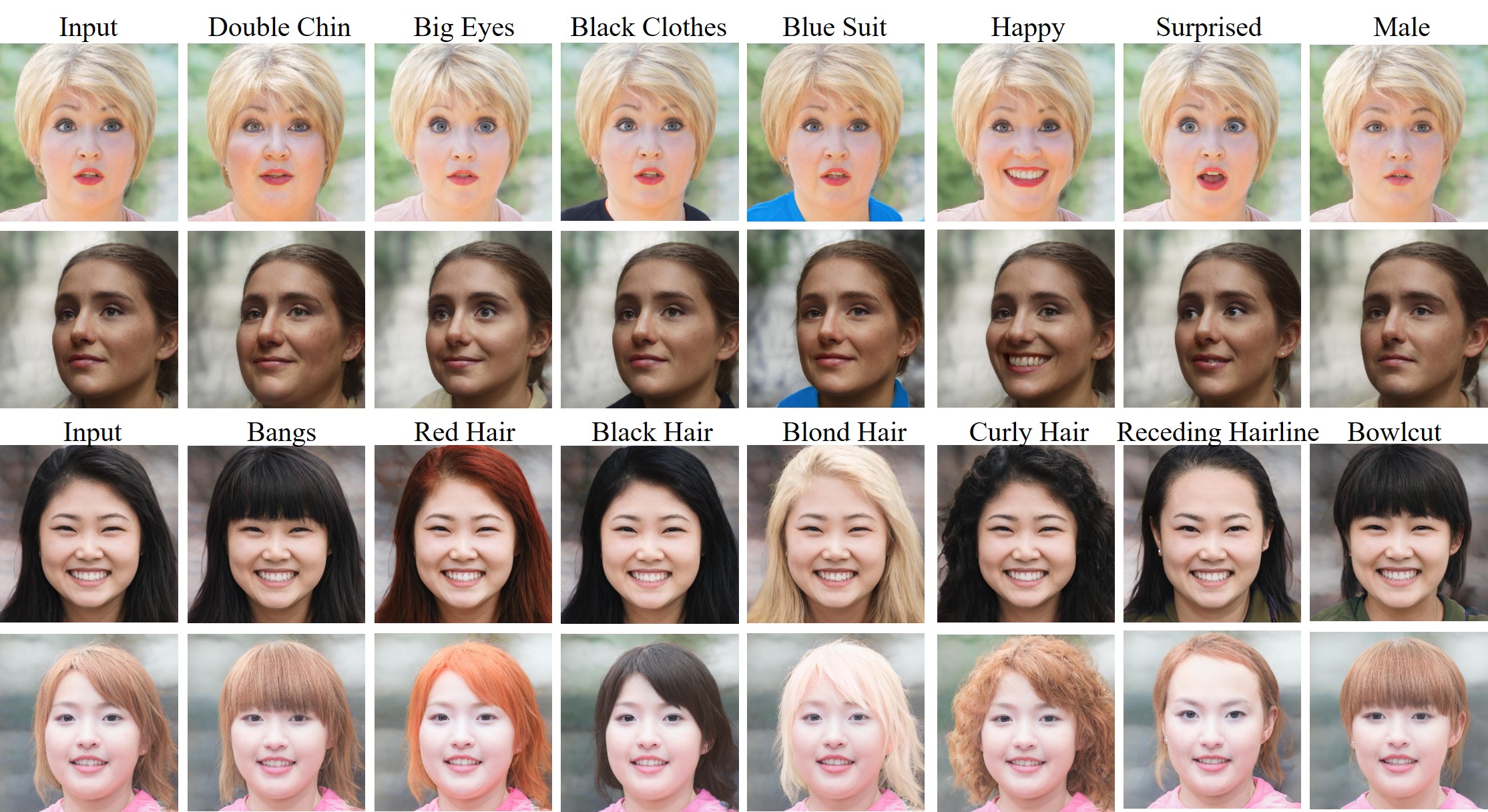}
	\end{center}
	\vspace{-3ex}
	\caption{Various edits for facial images on StyleGAN2 FFHQ model. The target attribute included in the text prompt is above each image. }
	\label{fig:face}
\end{figure*}

\begin{figure}[t]
	\begin{center}
		\includegraphics[width=0.85\linewidth]{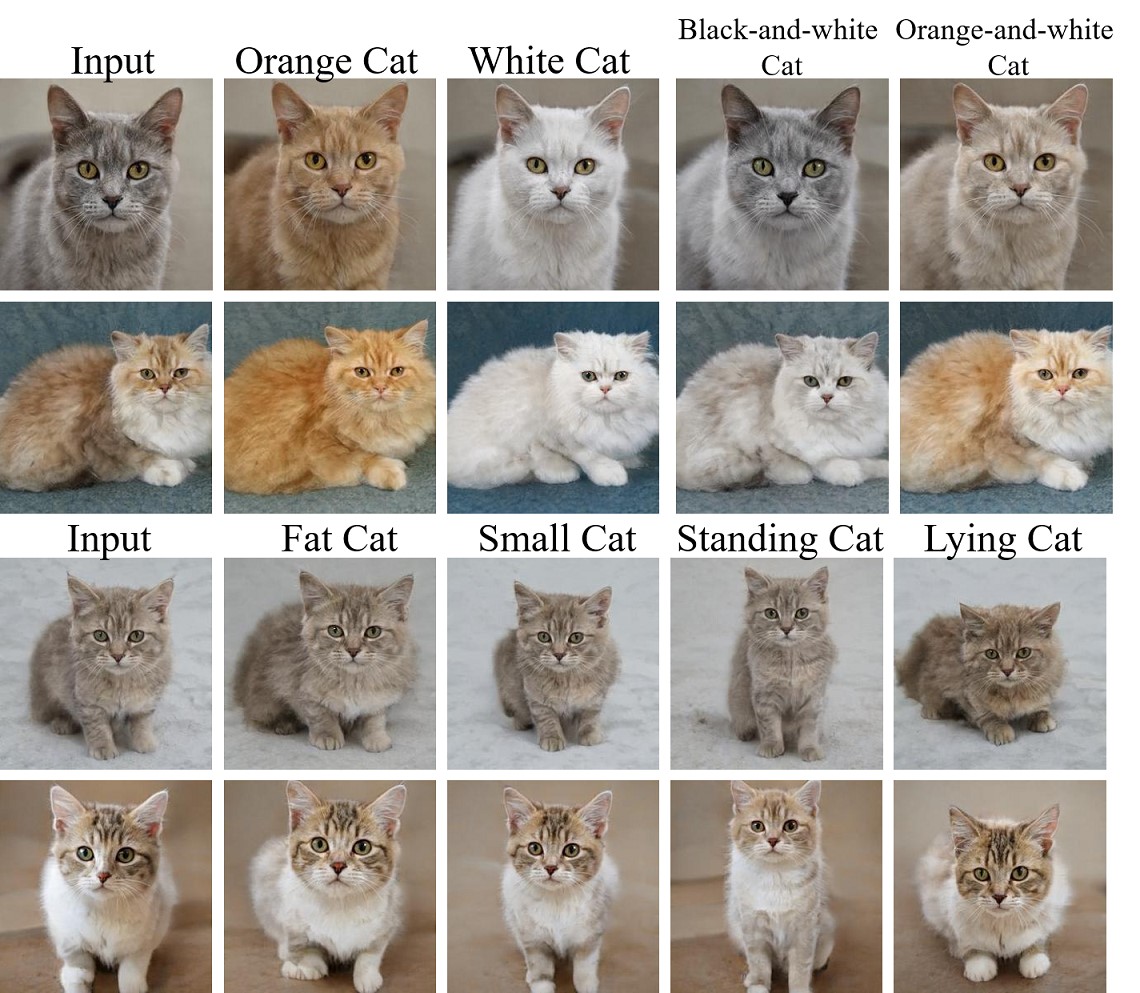}
	\end{center}
	\vspace{-0.25cm}
	\caption{A variety of edits for cat images, using StyleGAN2 pre-trained on LSUN cats dataset. The target text prompt is indicated above each column.}
	\label{fig:cat}
\end{figure}

\begin{figure*}[t]
	\begin{center}
		\includegraphics[width=0.95\linewidth]{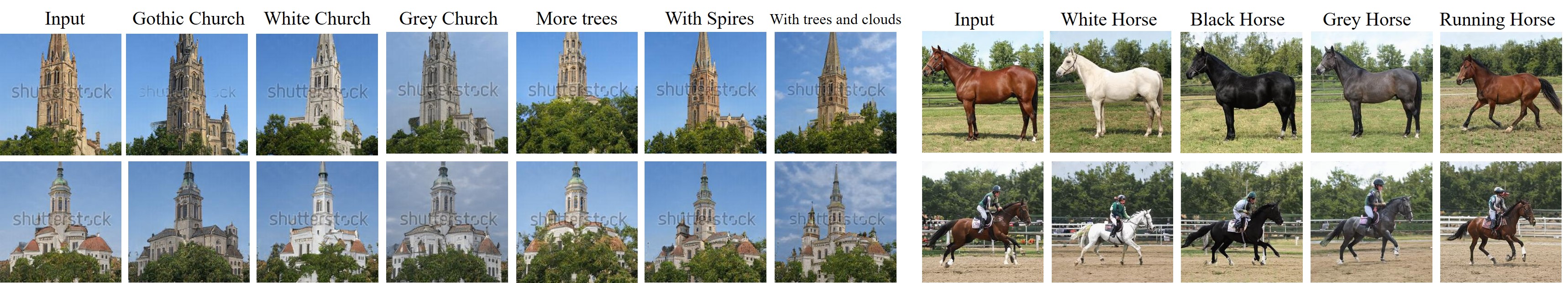}
	\end{center}
	\vspace{-3ex}
	\caption{Various edits for church and horse images, using StyleGAN2 pre-trained on LSUN churches dataset (left) and LSUN horses dataset (right) separately.}
	\label{fig:churchhorse}
\end{figure*}

\textbf{Various Text-driven Manipulation Results.}
To evaluate the generalization ability of our method that can be driven by texts with different semantic meanings, we show facial manipulation results of 21 attributes in Figure~\ref{fig:face}, which are generated from one trained model. The results show that only the target attributes are manipulated, while other irrelevant attributes are well preserved. Meanwhile, the results are well adapted to the individual with diverse details, rather than overfitting to the same color or shape, which can be seen in ``red hair'' and `` bowlcut hairstyle'' obviously. In addition, the manipulated results on images of cats, churches and horses are shown in Figure~\ref{fig:cat} and Figure~\ref{fig:churchhorse}, where StyleGAN2 pretrained on LSUN cat, church and horse datasets~\cite{yu2015lsun} are used.
It is worth mentioning that all text prompts have never been seen during training, which further indicates the effectiveness of the proposed method. 

\begin{figure}[t]
	\begin{center}
		\includegraphics[width=1\linewidth]{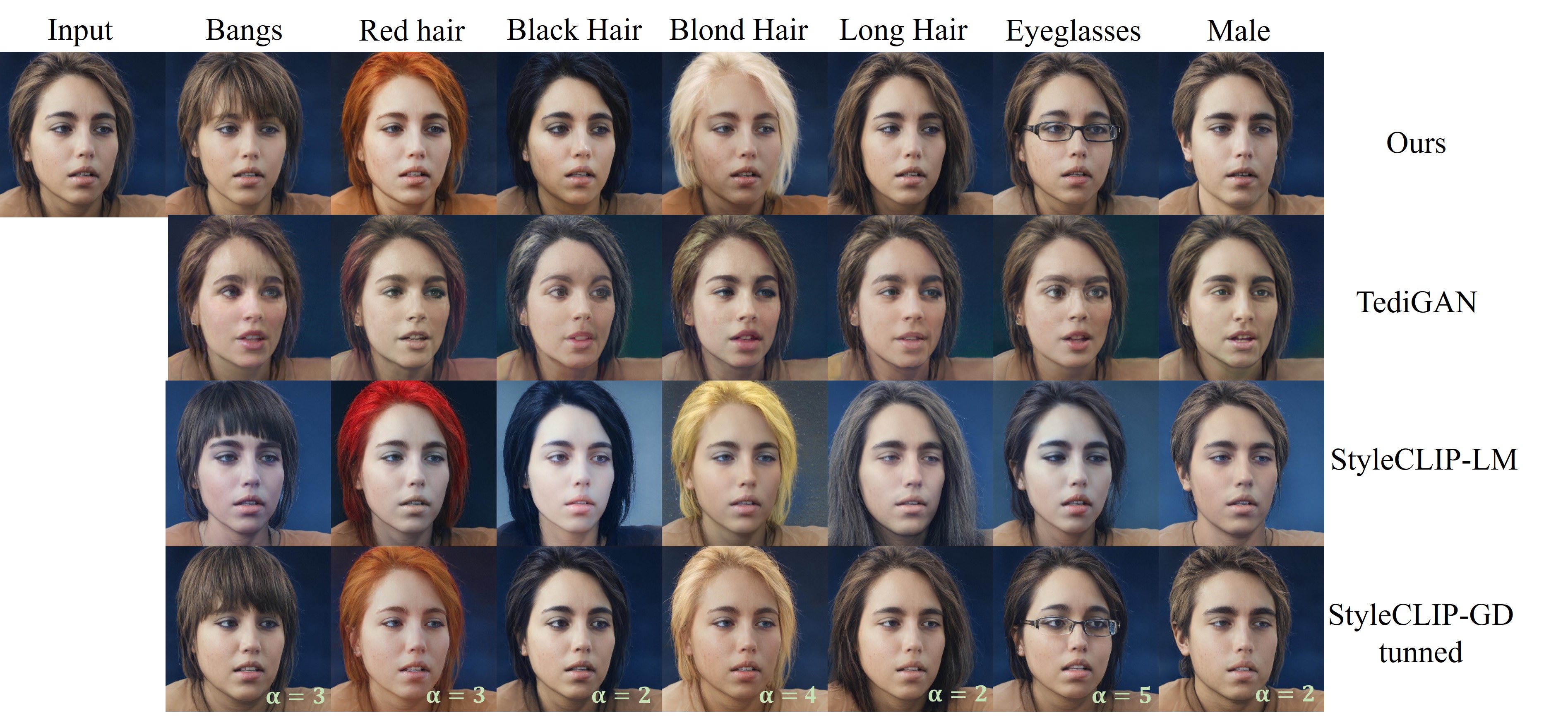}
	\end{center}
	\vspace{-3ex}
	\caption{Comparision results with TediGAN, StyleCLIP-LM and StyleCLIP-GD. Our approach demenstrates better visual realism and attribute disentanglement almost in each case.}
	\label{fig:comwm}
\end{figure}

\begin{figure}[t]
	\begin{center}
		\includegraphics[width=1\linewidth]{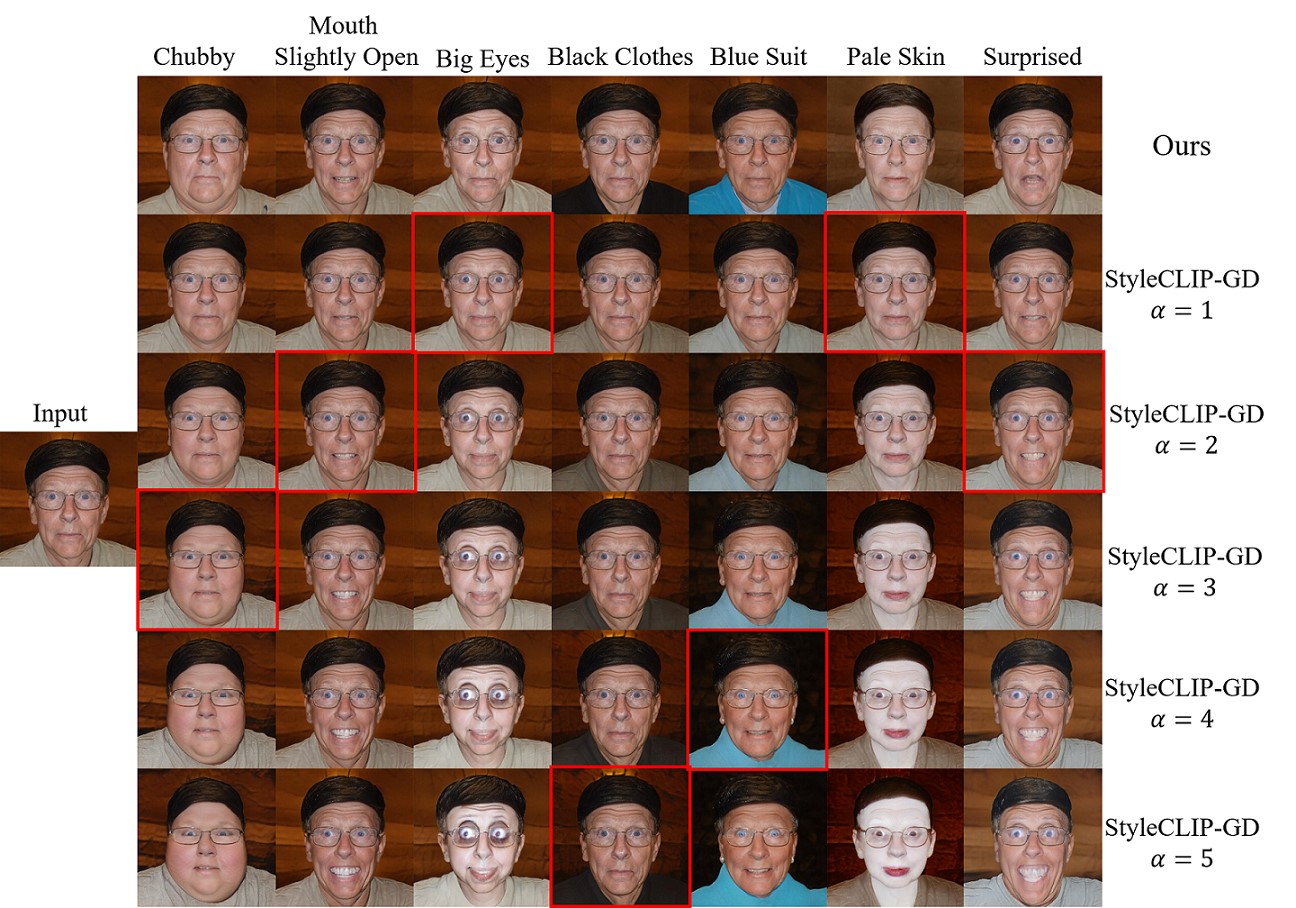}
	\end{center}
	\vspace{-3ex}
	\caption{The comparison between DeltaEdit and StyleCLIP-GD under different manipulation strength $\alpha$. Without complex hyper-parameters tuning, our method achieves more realistic and disentangled results, compared to the best results of StyleCLIP-GD under different settings~(labeled with red boxes).}
	\label{fig:com}
\end{figure}

\textbf{Comparisons.} 
We compare our method with state-of-the-art text-guided image manipulation methods, TediGAN~\cite{xia2021tedigan}, StyleCLIP-LM~\cite{patashnik2021styleclip}, and StyleCLI-GD~\cite{patashnik2021styleclip}. As shown in Figure~\ref{fig:comwm}, the results produced by TediGAN are highly entangled and almost fail in editing attributes, such as ``long hair'' and ``eyeglasses''. The results of StyleCLIP-LM are unstable and also entangled. Moreover, the results of StyleCLIP-GD have better disentanglement than StyleCLIP-LM as it is well-tunned for each case. However, the disadvantage of StyleCLIP-GD is the complex hyper-parameters tunning and high inference time reported in Table~\ref{table:time}. In comparison, our method yields the most impressive and disentangled results almost in each case.

To verify the inference flexibility of our method, we further compare our method with the strong baseline, \emph{i.e.}, StyleCLIP-GD, which achieves manipulation by manual tuning two hyper-parameters, disentanglement threshold $\beta$ and manipulation strength $\alpha$. Since we also introduce disentanglement threshold $\beta$ to control some channels of the obtained $\Delta s'$ as zero. For a fair comparison, we empirically set $\beta$ at 0.03 and compare with StyleCLIP-GD under different manipulation strength $\alpha$ in Figure~\ref{fig:com}. The results demonstrate that StyleGAN-GD cannot be generalized to different image attributes and produce feasible editing results under the same hyper-parameters. For example, the result conditioned on ``big eyes'' is the most accurate at $\alpha=1$ while the result conditioned on ``black clothes'' is the most correct when $\alpha=4$. In the contrast, our method achieves ideal manipulation results of various text prompts without manually tuning the manipulation strength. Moreover, under the same disentanglement threshold $\beta$, StyleCLIP-GD results tend to entangle with some irrelevant attributes. For example, for text prompt ``blue suit'', the generated results are entangled with ``blue eyes'', ``mouth slightly open'', ``wearing earrings'' and \emph{etc}. On the contrary, our method can achieve more disentangled results without complex hyper-parameters tuning.

\subsection{Quantitative Evaluation}
\hspace*{1em} In Table~\ref{tab:quatita}, we present objective measurements of FID, PSNR, and IDS (identity similarity before and after manipulation by Arcface) for comparison. The results are the average on ten editing texts. Compared with the state-of-the-art approaches, we achieve the best performance on all metrics. We also compare with the naïve solution to text-free training. For fairness, it is also trained in the $\mathcal{S}$ space and supervised by same loss functions in DeltaEdit. Compared with the naïve solution, the final solution (Ours) can largely improve the inference performance on all metrics, benefiting from the well-aligned CLIP delta image-text space. Finally, we conduct human subjective evaluations upon the manipulation accuracy (Acc) and visual realism (Real). We compare DeltaEdit with the strong baseline, StyleCLIP-GD, under different hyper-parameter $\alpha$ settings. In total 20 evaluation rounds are performed and 40 participants are invited. At each round, we present results of randomly sampled editing text to each participant. Participants were asked to choose the best text-driven image manipulation output considering Acc and to select the output images (not limited to 1 image) which are visually realistic (Real). The results are listed in Table \ref{tab:user}, showing the superiority of our model.
\begin{figure*}[t]
	\begin{center}
		\includegraphics[width=0.9\linewidth]{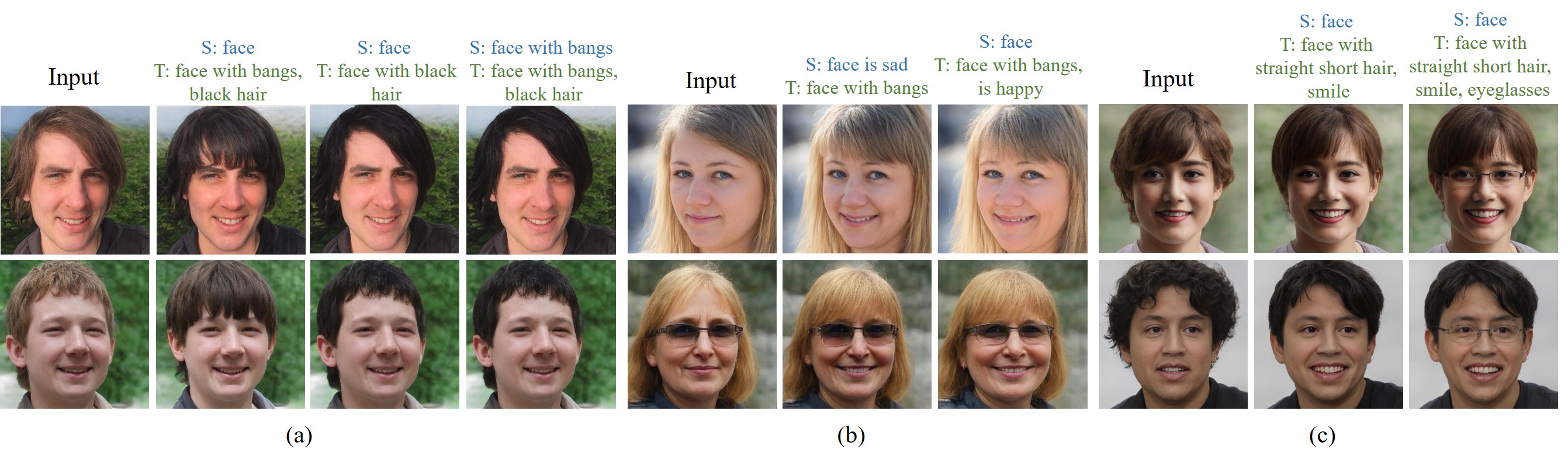}
	\end{center}
	\vspace{-5ex}
	\caption{Editing results under different text prompt settings, where the source text is labeled blue and the target text is labeled green. }
	\label{fig:prompt}
\end{figure*}

\subsection{Efficiency Analysis}
\hspace*{1em} To validate the efficiency and flexibility of the proposed method, we compare the computation time with TediGAN and three StyleCLIP approaches in Table~\ref{table:time}. Specifically, TediGAN first requires 12+ hours to encode images and texts into a common space and then trains the encoding module.
StyleCLIP-OP manipulates images with an optimization process, which requires several tens of seconds. StyleCLIP-LM is fast during inference, but needs 10-12 hours to train a mapper network for per text prompt. StyleCLIP-GD requires about 4 hours to pre-compute the global edit directions. However, since the pre-computed directions are rough and cannot be directly applied to control the manipulation, it requires additional manual tuning for different text prompts, which typically takes 8-9 seconds for each case. On the contrary, we avoid the high labor time during training and inference. Once trained, our Delta Mapper is universal and can directly manipulate images with new text prompt efficiently. Meanwhile, our method only needs 2-3 hours for training, since the training is directly conducted on the latent space without generating images in each iteration.

\begin{table}[t]
	\centering
    \resizebox{0.8\linewidth}{!}{
	\begin{tabular}{c|ccc}
	\toprule
    Methods & FID ($\downarrow$) & PSNR ($\uparrow$) &IDS ($\uparrow$) \\
    \midrule   
    TediGAN  &$31.13$ &$20.46$ &$0.60$\\
    StyleCLIP-LM  &$18.33$ &$21.41$ &$0.88$\\
    StyleCLIP-GD $\alpha=2$ &$12.06$ &$22.31$ &$0.86$\\
    StyleCLIP-GD $\alpha=3$ &$16.85$ &$19.50$ &$0.77$\\
    StyleCLIP-GD $\alpha=4$ &$22.85$ &$17.61$ &$0.68$\\
    \midrule
    Ours &$\mathbf{10.29}$ &$\mathbf{22.92}$ &$\mathbf{0.90}$\\
    Ours ($\mathcal{S}$ space) &$18.88$ &$13.49$ &$0.61$\\
    Ours ($\mathcal{W+}$ space) &$21.63$ &$12.83$ &$0.55$\\
    Ours (naïve) &$44.62$ &$11.21$ &$0.29$\\
    \bottomrule
    \end{tabular}}
    \caption{Quantitative comparison results on FID, PSNR, and IDS.}
	\label{tab:quatita}
\end{table}

\begin{table}[t]
	\centering
    \resizebox{0.8\linewidth}{!}{
	\begin{tabular}{c|cccc}
		\toprule
		& \makecell[c]{StyleCLIP \\$\alpha=2$} & \makecell[c]{StyleCLIP \\$\alpha=3$} & \makecell[c]{StyleCLIP \\$\alpha=4$} & Ours \\
		\midrule   
		Acc. ($\uparrow$) &$19\%$ & $9\%$ & $9.75\%$ & $\mathbf{62.25\%}$ \\
		\midrule
		Real. ($\uparrow$) &$50\%$ & $33\%$ & $28.75\%$ & $\mathbf{90.25}$\% \\
		\bottomrule
	\end{tabular}}
    \caption{User preference study on manipulation accuracy (Acc.) and visual realism (Real.)}
\label{tab:user}
\end{table}
\begin{table}[!tbp]
    \centering
    \resizebox{0.85\linewidth}{!}{
    \begin{tabular}{c|c|c|c|c|c}
	\toprule
    & Pre-proc. & Training time & Infer. time & \makecell[c]{Conditioned on \\input image} & \makecell[c]{Latent \\ space} \\
    \midrule   
    TediGAN &- & 12h+ & 20 s & yes & $\mathcal{W+}$ \\
    StyleCLIP-OP &- & - & 99 s & yes & $\mathcal{W+}$ \\
    StyleCLIP-LM &- & 10-12h & 70ms & yes & $\mathcal{W+}$ \\
    StyleCLIP-GD &4h & - & 72ms+$T_{labor}$ & no & $\mathcal{S}$ \\
    Ours &4h & 2.7h & 73ms & yes & $\mathcal{S}$ \\
	\bottomrule
    \end{tabular}}
    \caption{The time comparison of our method with other state-of-the-art methods. For StyleCLIP-GD, the additional time of manually hyper-parameters tuning during the inference is denoted as $T_{labor}$, which is typically 8-9 seconds for each case.}
	\label{table:time}
\end{table}

\subsection{Text-prompts Analysis}
\hspace*{1em} In Figure~\ref{fig:prompt}, we explore how different text prompt settings can affect the editing results. During inference, the editing direction $\Delta s'$ is driven by the CLIP text space direction $\Delta t$ between source text and target text.
(a) We first construct three different (source text, target text) pairs and find that regardless of the content of them, the editing results are affected by the difference between them. For example, although both the source and target texts contain the attribute of ``bangs'', the editing result is only influenced by the difference, namely ``black hair''.
(b) Moreover, to align with the training phase, we construct (source text, target text) pairs are each a specific text description for an image, such as “face is sad” as the source text and “face with bangs” as the target text. We find that the generated face to become more ``happy'' (the opposite direction of the source text) and with ``bangs'' (the direction of the target text), which is equivalent to directly putting the attributes to be edited all into the target text.
(c) In addition, we construct target texts with multiple combinations of facial attributes, including hairstyles, smile and eyeglasses, and our method can yield desired results driven by text prompts containing multiple semantics. Note that our method can directly perform multi-attributes manipulation without additional training processes, since the different manipulation direction has been learned well by training on large-scale data.
\emph{More exploration experiments about text prompts are seen in Supp.}

\subsection{Ablation Study}

\begin{figure}[t]
	\begin{center}
	\includegraphics[width=0.8\linewidth]{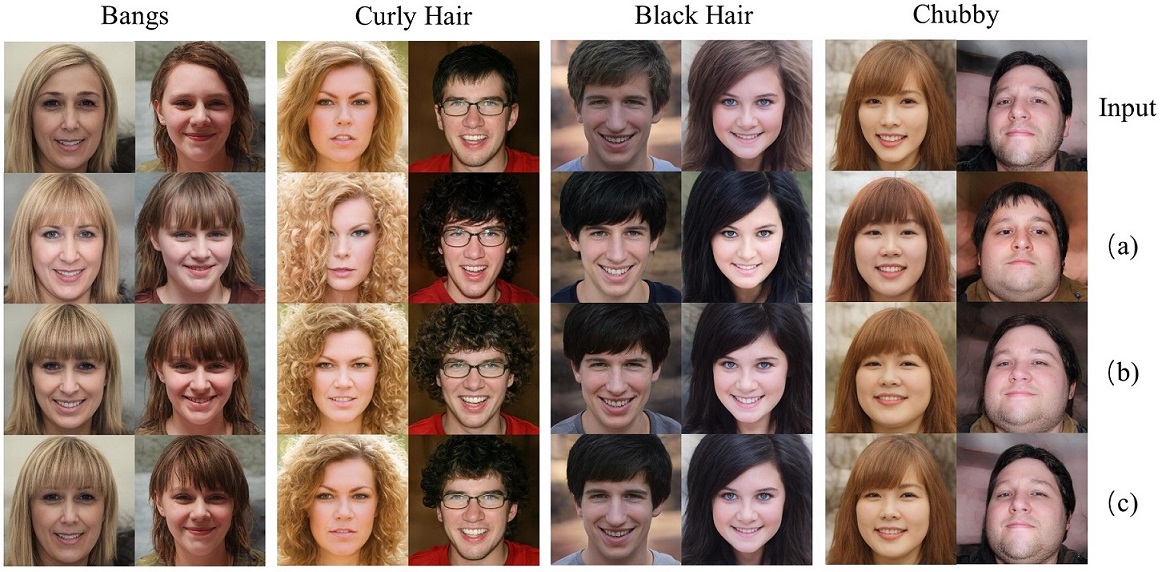}
	\end{center}
	\vspace{-3ex}
	\caption{Manipulation results by implementing our DeltaEdit in (a) $\mathcal{W+}$ space, (b) $\mathcal{S}$ space, and (c) $\mathcal{S}$ space with relevance matrix $R_s$.}
	\label{fig:ablation}
\end{figure}

\hspace*{1em} \textbf{Choice of the manipulation space.}
To find an appropriate space for the proposed method, we conduct experiments by performing manipulations in $\mathcal{W+}$ space and $\mathcal{S}$ space, respectively. 
The quantitative and qualitative results in Table~\ref{tab:quatita} and Figure~\ref{fig:ablation} shows that $\mathcal{S}$ results can achieve better visual quality and identity preservation than $\mathcal{W+}$. Thus, we implement our DeltaEdit in $\mathcal{S}$ space, and introduce $R_s$ to further improve the disentanglement performance.

\textbf{The effectiveness of relevance matrix $R_s$.} For further improving the disentanglement, we introduce relevance matrix $R_s$ to limit some irrelevant channels from changing. The fourth row in Figure~\ref{fig:ablation} shows that, with $R_s$, our method can successfully edit the desired attributes while preserving the text-irrelevant content unchanged. For example, for "bangs", the generated results have bangs while unchanging the background or facial pose.

\section{Conclusions}
\hspace*{1em} In this paper, we propose DeltaEdit to support various edits in one single model without the need of training many independent models or complex manual tuning. It is trained in a text-free manner and gets rid of an expensive collection of image-text pairs. Once trained, it can be well generalized to any unseen text prompt for zero-shot inference. Extensive qualitative and quantitative experiments demonstrate the superiority of our method, in terms of the capability of generating high-quality results, the efficiency in the training or inference phase, and the generalization to any unseen text prompt.

\textbf{Acknowledgement.} 
This work was supported by National Key Research and Development Program of China under Grant No. 2021YFC3320103.

{
    \small
    \bibliographystyle{ieeenat_fullname}
    \bibliography{output}
}


\end{document}